\documentclass[11pt, a4paper, logo, twocolumn, copyright]{googledeepmind}

\usepackage[authoryear, sort&compress, round]{natbib}
\usepackage[textwidth=18mm]{todonotes}
\usepackage{graphicx}
\usepackage{makecell}
\usepackage{booktabs}
\usepackage{array}
\usepackage{multirow}
\usepackage{tikz}
\usepackage{xcolor}
\usetikzlibrary{shapes.arrows, positioning, shadows, calc}

\definecolor{gdm_darkgray}{HTML}{9aa0a6}
\definecolor{gdm_gray}{HTML}{3c4043}

\newcommand\blfootnote[1]{%
  \begingroup
  \renewcommand\thefootnote{}\footnote{#1}%
  \addtocounter{footnote}{-1}%
  \endgroup
}

\renewcommand{\today}{2025-03-12}

\newcommand{\wmtpp}{WMT24\nolinebreak\hspace{-.05em}\raisebox{.4ex}{\tiny\bf+}\nolinebreak\hspace{-.10em}\raisebox{.4ex}{\tiny\bf +}}

\title{Gemma 3 Technical Report}

\author{Gemma Team, Google DeepMind\authfootnotemark{1}}

\authfootnotetext{1}{See Contributions and Acknowledgments section for full author list. Please send correspondence to {\tt gemma-3-report@google.com}.}

\begin{abstract}
We introduce Gemma~3, a multimodal addition to the Gemma family of lightweight open models, ranging in scale from 1 to 27 billion parameters. 
This version introduces vision understanding abilities, a wider coverage of languages and longer context -- at least 128K tokens.
We also change the architecture of the model to reduce the KV-cache memory that tends to explode with long context. 
This is achieved by increasing the ratio of local to global attention layers, and keeping the span on local attention short. 
The Gemma~3 models are trained with distillation and achieve superior performance to Gemma~2 for both pre-trained and instruction finetuned versions.
In particular, our novel post-training recipe significantly improves the math, chat, instruction-following and multilingual abilities, making Gemma3-4B-IT competitive with Gemma2-27B-IT and Gemma3-27B-IT comparable to Gemini-1.5-Pro across benchmarks. 
We release all our models to the community.
\end{abstract}

\begin{document}

\maketitle

\section{Introduction}

We present the newest version of Gemma open language models~\citep{gemmateam2024gemma}, co-designed with the family of Gemini frontier models~\citep{geminiteam2023gemini}.
This new version comes in sizes comparable to Gemma~2~\citep{gemmateam2024gemma2}, with the addition of a 1B model. 
These models are designed to run on standard consumer-grade hardware such as phones, laptops, and high-end GPUs. This version comes with several new abilities to the Gemma family; namely, multimodality, long context, and multilinguality, while preserving or surpassing the performance of prior versions. 

In terms of multimodality, most Gemma~3 models are compatible with a tailored version of the SigLIP vision encoder~\citep{zhai2023sigmoid}.
The language models treat images as a sequence of soft tokens encoded by SigLIP. We reduce the inference cost of image processing by condensing the vision embeddings into a fixed size of 256 vectors.
The encoder works at a fixed resolution and we take inspiration from LLaVA~\citep{liu2024visual} to enable flexible resolutions with a Pan and Scan (P\&S) method. 

The second main architectural improvement is an increase in context size to 128K tokens, without reducing performance.
A challenge with long context is the memory explosion of the KV cache during inference.
To reduce this issue, we interleave multiple local layers between each global layer, and assign a smaller span of only 1024 tokens to the local layers.
Therefore, only the global layers attend to long context, and we have 1 global for every 5 local layers.

The pre-training optimization recipe is similar to Gemma~2, with some modifications in the architecture design.
We use the same tokenizer as Gemini 2.0, and we also revisit our data mixture to improve the multilingual capabilities of the models, while introducing image understanding.
All Gemma 3 models are trained with knowledge distillation~\citep{hinton2015distilling}.

In post-training, we focus our efforts on improving mathematics, reasoning, and chat abilities, as well as integrating the new capabilities of Gemma~3, long-context, and image inputs.
We use a novel post-training approach that brings gains across all capabilities, including math, coding, chat, instruction following, and multilingual. The resulting Gemma~3 instruction-tuned models are both powerful and versatile, outperforming their predecessors by a wide margin.

In the following sections, we provide a brief overview of our models, including the architecture and pre- and post-training recipes. 
We also provide detailed evaluations across a wide variety of quantitative and qualitative benchmarks. We discuss our approach to safe and responsible deployment and outline the broader implications of Gemma~3, its limitations, and advantages.

\begin{figure}[t!]
    \centering
    \includegraphics[width=.6\linewidth]{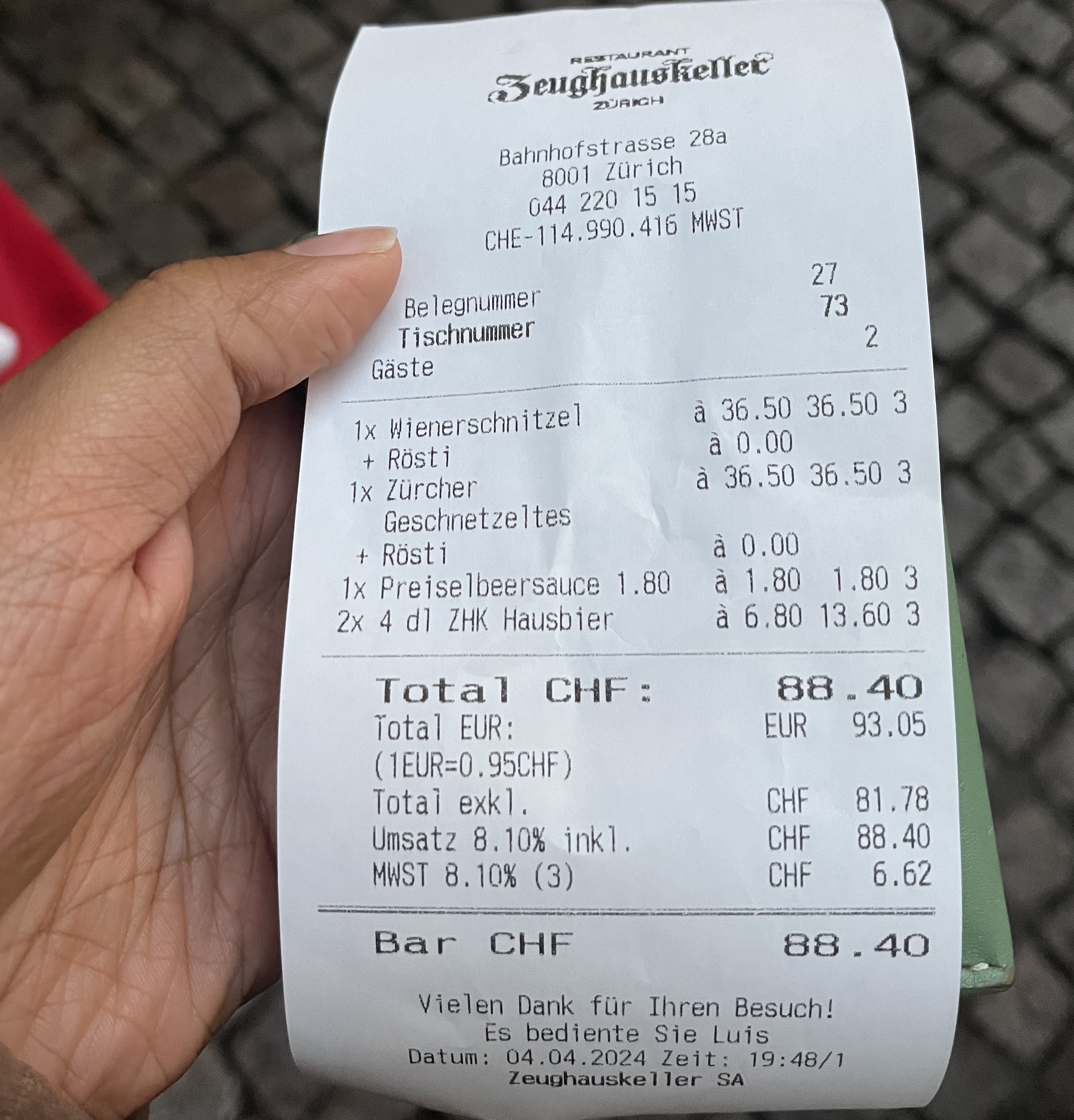}
    \includegraphics[width=\linewidth]{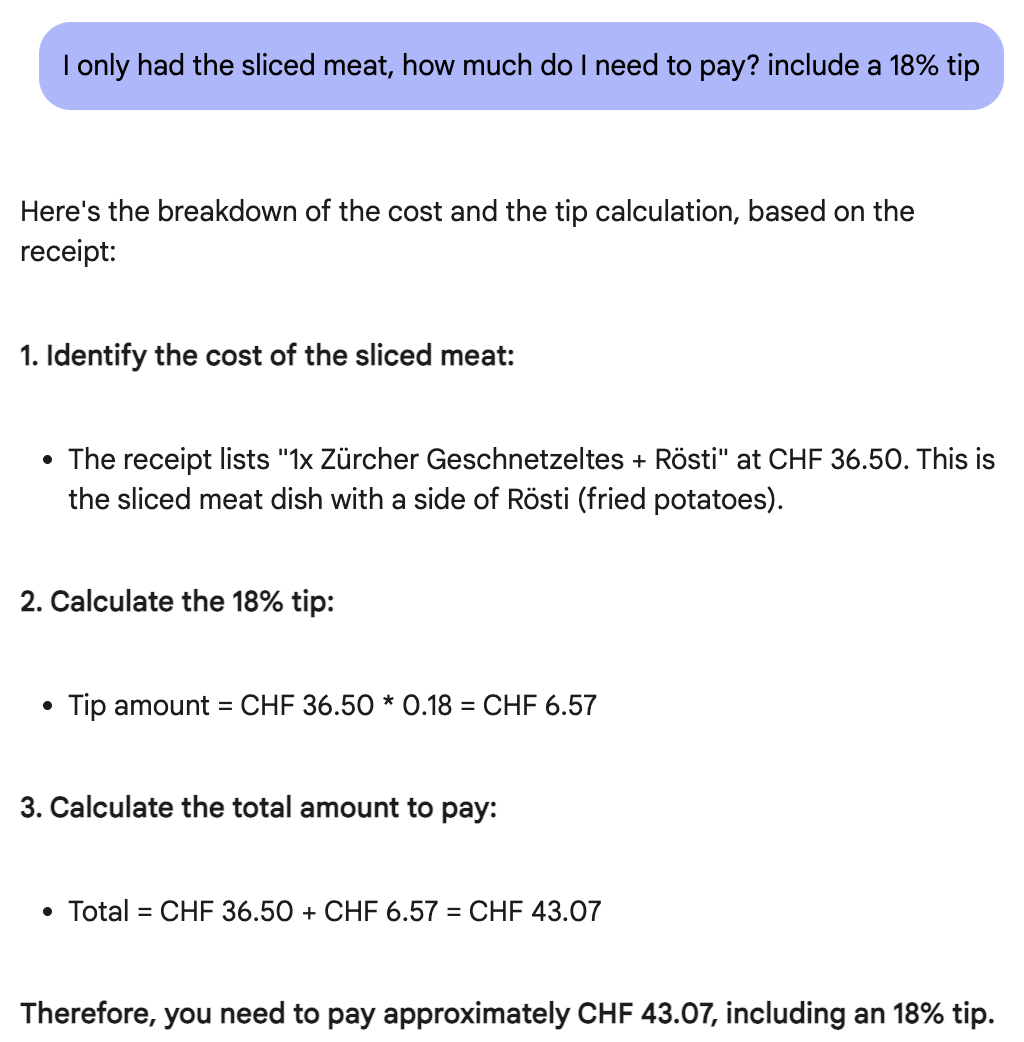}
    \caption{Example of visual interaction with Gemma 3 27B IT model.}
    \label{fig:mm_example}
\end{figure}

\section{Model Architecture}
\label{sec:architecture}

Gemma~3 models follow the same general decoder-only transformer architecture as previous iterations~\citep{DBLP:journals/corr/VaswaniSPUJGKP17}, with most architecture elements similar to the first two Gemma versions. 
We use a Grouped-Query Attention~(GQA)~\citep{ainslie2023gqa} with post-norm and pre-norm with RMSNorm~\citep{rmsnorm}.
Inspired by~\cite{dehghani2023scaling}, \cite{wortsman2023small} and \cite{team2024chameleon}, we replace the soft-capping of Gemma~2 with QK-norm.
In this section, we focus on some key differences from previous versions below.

\noindent\textbf{5:1 interleaving of local/global layers.} 
We alternate between a local sliding window self-attention~\citep{beltagy2020longformer} and global self-attention~\citep{globalattention}, with a pattern of 5 local layers for every global layer, starting with a local layer as the first layer of the model.

\begin{table}[t]
    \centering
    \begin{tabular}{@{}l r r r@{}}
    \toprule
        Model & \makecell{Vision\\ Encoder} & \makecell{Embedding\\Parameters} & \makecell{Non-embedding\\Parameters} \\
        \midrule
        \textbf{1B}  & 0 & 302M & 698M \\
        \textbf{4B}  & 417M & 675M & 3,209M \\
        \textbf{12B} & 417M & 1,012M & 10,759M \\
        \textbf{27B} & 417M & 1,416M & 25,600M \\
    \bottomrule
    \end{tabular}
    \caption{Parameter counts for the Gemma~3 models. 
    Our vocabulary has 256k entries.}
    \label{tab:model_param_counts}
\end{table}

\noindent\textbf{Long context.}
Gemma~3 models support context length of 128K tokens, with the exception of the 1B model that has 32K. We increase RoPE base frequency from 10k to 1M on global self-attention layers, and keep the frequency of the local layers at 10k. We follow a process similar to the positional interpolation of \cite{chen2023extending} to extend the span of the global self-attention layers.

\subsection{Vision modality}

\noindent\textbf{Vision encoder.}
We use a 400M variant of the SigLIP encoder \citep{zhai2023sigmoid}, a Vision Transformer~\citep{dosovitskiy2020image} trained with a variation of the CLIP loss~\citep{radford2021learning}.
The Gemma vision encoder takes as input square images resized to 896 x 896, and is finetuned on data from visual assistant tasks.
For simplicity, we share the vision encoder across our 4B, 12B, and 27B models, keeping it frozen during training. 

\noindent\textbf{Pan \& Scan (P\&S).} The Gemma vision encoder operates at a fixed resolution of 896 $\times$ 896. This results in artifacts when processing non-square aspect ratios and high-resolution images, leading to unreadable text, or small objects disappearing.
We address this issue with an adaptive windowing algorithm during inference. This algorithm segments images into non-overlapping crops of equal size, covering the whole image, and resize them to 896×896 pixels to pass them to the encoder. 
This windowing is applied only when necessary, and control for the maximum number of crops. 
It is an inference-time only optimization and can be disabled for faster inference. 

\subsection{Pre-training}

We follow a similar recipe as in Gemma~2 for pre-training with knowledge distillation.

\noindent\textbf{Training data.}
We pre-train our models on a slightly larger token budget than Gemma~2, i.e., we train on 14T tokens for Gemma~3 27B, 12T for the 12B version, 4T for the 4B, and 2T tokens for the 1B. 
The increase in tokens accounts for the mix of images and text used during pre-training.
We also increase the amount of multilingual data to improve language coverage.
We add both monolingual and parallel data, and we handle the imbalance in language representation using a strategy inspired by \cite{chung2023unimaxfairereffectivelanguage}. 

\noindent\textbf{Tokenizer.}
We use the same tokenizer as Gemini 2.0: a SentencePiece tokenizer with split digits, preserved whitespace, and byte-level encodings~\citep{kudo-richardson-2018-sentencepiece}.
The resulting vocabulary has 262k entries.
This tokenizer is more balanced for non-English languages.

\begin{table}[t]
 \setlength{\tabcolsep}{5pt}
    \centering
    \begin{tabular}{@{}l c c c c c@{}}
    \toprule
    & & & \multicolumn{3}{c}{Shards} \\
    \cmidrule{4-6}
    Model & Type & \#Chips & Data & Seq. & Replica \\
    \midrule
        \textbf{1B} & TPUv5e & 512 & 16 & 16 & 2 \\
        \textbf{4B} & TPUv5e & 2048 & 16 & 16 & 8 \\
        \textbf{12B} & TPUv4 & 6144 & 16 & 16 & 24 \\
        \textbf{27B} & TPUv5p & 6144 & 24 & 8 & 32 \\
    \bottomrule
    \end{tabular}
    \caption{Training infrastructure with sharding by data, sequence (Seq.), and replica.}
    \label{tab:training_infra_sharding}
\end{table}
 
\noindent\textbf{Filtering.}
We use filtering techniques that reduce the risk of unwanted or unsafe utterances and remove certain personal information and other sensitive data.
We decontaminate evaluation sets from our pre-training data mixture, and reduce the risk of recitation by minimizing the proliferation of sensitive outputs. 
We also apply a quality reweighing step inspired by~\cite{sachdeva2024train} to reduce occurrences of low quality data.

\noindent\textbf{Distillation.}
We sample 256 logits per token, weighted by teacher probabilities. The student learns the teacher's distribution within these samples via cross-entropy loss. The teacher's target distribution is set to zero probability for non-sampled logits, and renormalized.

\subsection{Quantization Aware Training}

Along with the raw checkpoints, we also provide quantized versions of our models in different standard formats. 
These versions are obtained by finetuning each model for a small number of steps, typically 5,000, using Quantization Aware Training (QAT)~\citep{jacob2018quantization}.
We use probabilities from the non-quantized checkpoint as targets, and adapt the data to match the pre-training and post-training distributions. Based on the most popular open source quantization inference engines (e.g. llama.cpp), we focus on three weight representations: per-channel int4, per-block int4, and switched fp8.
In Table~\ref{tab:quantization}, we report the memory filled by raw and quantized models for each weight representation with and without a KV-cache for a sequence of 32k tokens.

\begin{table}[t]
 \setlength{\tabcolsep}{5pt}
    \centering
    \begin{tabular}{@{}l c c c c@{}}
    \toprule
    & Raw (GB) & \multicolumn{3}{c}{Quantized (GB)} \\
    \cmidrule{2-5}
    Model & bf16 & Int4 & $\text{Int4}_{\text{blocks=32}}$ & SFP8 \\
    \midrule
        \textbf{1B} & 2.0 & 0.5 & 0.7 & 1.0 \\
        +KV & 2.9 & 1.4 & 1.6 & 1.9 \\
    \cmidrule{2-5}
        \textbf{4B} & 8.0 & 2.6 & 2.9 & 4.4  \\
        +KV & 12.7 & 7.3 & 7.6 & 9.1 \\
    \cmidrule{2-5}
        \textbf{12B} & 24.0 & 6.6 & 7.1 & 12.4 \\
        +KV & 38.9 & 21.5 & 22.0 & 27.3 \\
    \cmidrule{2-5}
        \textbf{27B} & 54.0 & 14.1 & 15.3 & 27.4 \\
        +KV & 72.7 & 32.8 & 34.0 & 46.1 \\
    \bottomrule
    \end{tabular}
    \caption{Memory footprints (in GB) comparison between raw (bfloat16) and quantized checkpoints for weights and KV caching (+KV) at 32,768 context size, quantized in 8 bits.}
    \label{tab:quantization}
\end{table}

\subsection{Compute Infrastructure}

We train our models with TPUv4, TPUv5e, and TPUv5p as outlined in Table \ref{tab:training_infra_sharding}.
Each model configuration is optimized to minimize training step time. 
For the vision encoder, we pre-compute the embeddings for each image and directly train with the embeddings, adding no cost to the training of the language models.  

The optimizer state is sharded using an implementation of ZeRO-3~\citep{ren2021zero}. 
For multi-pod training, we perform a data replica reduction over the data center network, using the Pathways approach of \cite{barham2022pathways}.
We use the `single controller' programming paradigm of Jax \citep{bradburyJAX} and Pathways \citep{barham2022pathways}, along with the GSPMD partitioner~\citep{gspmd} and the MegaScale XLA compiler~\citep{xla}.

\begin{table}[t]
    \centering
    \footnotesize
    \begin{tabular}{p{0.3\linewidth} p{0.4\linewidth}}
    \toprule
    \textbf{Context} & \textbf{Formatting} \\
        \midrule
        User turn & \texttt{\color{NavyBlue}<start\_of\_turn>user} \\
        \midrule
        Model turn & \texttt{\color{NavyBlue}<start\_of\_turn>model} \\
        \midrule
        End of turn & \texttt{\color{NavyBlue}<end\_of\_turn>} \\
         \midrule
    \multicolumn{2}{c}{\textbf{Example of discussion:}} \\
    \midrule
   \multicolumn{2}{p{0.8\linewidth}}{
        \textbf{User:} \texttt{Who are you?}\par
        \textbf{Model:} \texttt{My name is Gemma!}\par
        \textbf{User:} \texttt{What is 2+2?}\par
        \textbf{Model:} \texttt{2+2=4.}
    } \\
    \midrule
    \multicolumn{2}{c}{\textbf{Model input:}} \\
    \midrule
   \multicolumn{2}{p{0.8\linewidth}}{
        \texttt{\color{red}[BOS]\color{NavyBlue}<start\_of\_turn>user}\par
        \texttt{Who are you?}\texttt{\color{NavyBlue}<end\_of\_turn>}\par
        \texttt{\color{NavyBlue}<start\_of\_turn>model}\par
        \texttt{My name is Gemma!\color{NavyBlue}<end\_of\_turn>}\par
        \texttt{\color{NavyBlue}<start\_of\_turn>user}\par
        \texttt{What is 2+2?}\texttt{\color{NavyBlue}<end\_of\_turn>}\par
        \texttt{\color{NavyBlue}<start\_of\_turn>model}
    } \\
    \midrule
    \multicolumn{2}{c}{\textbf{Model output:}} \\
    \midrule
   \multicolumn{2}{p{0.8\linewidth}}{
        \texttt{2+2=4.}\texttt{\color{NavyBlue}<end\_of\_turn>}
    } \\
    \bottomrule
    \end{tabular}
    \caption{Formatting for Gemma IT models. Explicitly add the \texttt{\color{red}[BOS]} token after tokenization, or use the \texttt{add\_bos=True} option in the tokenizer. \textit{Do not tokenize the text "[BOS]"}.}
    \label{tab:formatting_tokens}
    \vspace{-0.5cm}
\end{table}

\begin{table*}[t]
\footnotesize
\centering
\begin{tabular}{llccccc}
\toprule
Rank & Model & Elo   & 95\% CI        & Open       &Type & \#params/\#activated\\
\midrule
1       & Grok-3-Preview-02-24  & 1412  & +8/-10        & - & - & -\\
1       & GPT-4.5-Preview       & 1411  & +11/-11       & -& - & -\\
3       & Gemini-2.0-Flash-Thinking-Exp-01-21   & 1384  & +6/-5 & - & - & -\\
3       & Gemini-2.0-Pro-Exp-02-05      & 1380  & +5/-6 & - & - & -\\
3       & ChatGPT-4o-latest (2025-01-29)        & 1377  & +5/-4 & - & - & -\\
6       & DeepSeek-R1   & 1363  & +8/-6 & yes & MoE & 671B/37B\\
6       & Gemini-2.0-Flash-001  & 1357  & +6/-5 & - & - & -\\
8       & o1-2024-12-17 & 1352  & +4/-6 & - & - & -\\
\color{blue!70!black} 9      & \color{blue!70!black} \textbf{Gemma-3-27B-IT}        & \color{blue!70!black}\textbf{1338}  & \color{blue!70!black} \textbf{+8/-9} & \color{blue!70!black} \textbf{yes} & \color{blue!70!black} \textbf{Dense} & \color{blue!70!black} \textbf{27B}\\
9      & Qwen2.5-Max   & 1336  & +7/-5 & - & - & -\\
9      & o1-preview    & 1335  & +4/-3 & - & - & -\\
9      & o3-mini-high  & 1329  & +8/-6 & - & - & -\\
13      & DeepSeek-V3   & 1318  & +8/-6 & yes & MoE & 671B/37B\\
14      & GLM-4-Plus-0111       & 1311  & +8/-8 & - & - & -\\
14      & Qwen-Plus-0125        & 1310  & +7/-5 & - & - & -\\
14      & Claude 3.7 Sonnet     & 1309  & +9/-11 & - & - & -\\
14      & Gemini-2.0-Flash-Lite   & 1308  & +5/-5 & - & - & -\\
18      & Step-2-16K-Exp        & 1305  & +7/-6 & - & - & -\\
18      & o3-mini       & 1304  & +5/-4 & - & - & -\\
18      & o1-mini       & 1304  & +4/-3 & - & - & -\\
18      & Gemini-1.5-Pro-002    & 1302  & +3/-3 & - & - & -\\
...     &       &       &       &       &       & \\
28      & Meta-Llama-3.1-405B-Instruct-bf16     & 1269  & +4/-3 & yes & Dense & 405B\\
...     &       &       &       &       &       & \\
38      & Llama-3.3-70B-Instruct        & 1257  & +5/-3 & yes & Dense & 70B\\
...     &       &       &       &       &       & \\
39      & Qwen2.5-72B-Instruct  & 1257  & +3/-3 & yes & Dense & 72B\\
...     &       &       &       &       &       & \\
59      & Gemma-2-27B-it        & 1220  & +3/-2 & yes & Dense & 27B\\
\bottomrule
\end{tabular}
\caption{Evaluation of Gemma~3 27B IT model in the Chatbot Arena \citep{chiang2024chatbot}. All the models are evaluated against each other through blind side-by-side evaluations by human raters. Each model is attributed a score, based on the Elo rating system.
\emph{Gemma-3-27B-IT numbers are preliminary results received on March 8, 2025}.}
 \label{tab:lmsys_elo_leaderboard}
\end{table*}

\section{Instruction-Tuning}
Pre-trained models are turned into instruction-tuned models with an improved post-training approach compared to our prior recipe (see Table~\ref{tab:it_fs}).

\noindent\textbf{Techniques.}
Our post-training approach relies on an improved version of knowledge distillation~\citep{hinton2015distilling,anil2018large,agarwal2024policy} from a large IT teacher, along with a RL finetuning phase based on improved versions of BOND~\citep{sessa2024bondaligningllmsbestofn}, WARM~\citep{rame2024warm}, and WARP~\citep{rame2024warp}.

\noindent\textbf{Reinforcement learning objectives.} We use a variety of reward functions to improve helpfulness, math, coding, reasoning, instruction-following, and multilingual abilities, while minimizing model harmfulness. This includes learning from weight averaged reward models~\citep{rame2024warm} trained with human feedback data, code execution feedback~\citep{gehring2024rlef}, and ground-truth rewards for solving math problems~\citep{lambert2024tulu,deepseekai2025deepseekr1}.

\noindent\textbf{Data filtering.} We carefully optimize the data used in post-training to maximize model performance. We filter examples that show certain personal information, unsafe or toxic model outputs, mistaken self-identification data, and duplicated examples. Including subsets of data that encourage better in-context attribution, hedging, and refusals to minimize hallucinations also improves performance on factuality metrics, without degrading model performance on other metrics. 

\noindent\textbf{[BOS] token.} For both PT and IT models, text starts with a \texttt{\color{red}[BOS]} token, that needs to be added explicitly since the text ``[BOS]'' does not map to the \texttt{\color{red}[BOS]} token.
For instance, Flax has an option, \texttt{add\_bos=True}, to add this token automatically when tokenizing.
An example of the formatting for an IT model is shown in Table~\ref{tab:formatting_tokens}, 

\noindent\textbf{PT versus IT Formatting.} 
All models share the same tokenizer, with some control tokens dedicated to IT formatting.
A key difference is that PT models output a \texttt{<eos>} token at the end of generation, while IT models output a \texttt{<end\_of\_turn>}  at the end of the generation, as shown for IT in Table~\ref{tab:formatting_tokens}.
Fine-tuning either model type thus also requires adding their respective end tokens.

\phantomsection
\section{Evaluation of final models}
\label{sec:evals}

In this section, we evaluate the IT models over a series of automated benchmarks and human evaluations across a variety of domains, as well as static benchmarks such as MMLU.

\subsection{LMSYS Chatbot Arena}

In this section, we report the performance of our IT 27B model on LMSys Chatbot Arena \citep{chiang2024chatbot} in blind side-by-side evaluations by human raters against other state-of-the-art models. 
We report Elo scores in Table \ref{tab:lmsys_elo_leaderboard}. 
Gemma~3 27B IT (1338) is among the top 10 best models, with a score above other non-thinking open models, such as DeepSeek-V3 (1318), LLaMA~3 405B (1257), and Qwen2.5-70B (1257), which are much larger models.
Finally, the Elo of Gemma~3 is significantly higher than Gemma~2, at 1220.
Note that Elo scores do not take into account visual abilities, which none of the aforementioned models have.

\begin{table*}[th!]
\footnotesize
\centering
\begin{tabular}{@{}l cc c cc c ccc c cccc @{}}
\toprule
& \multicolumn{2}{c}{Gemini 1.5} && \multicolumn{2}{c}{Gemini 2.0} && \multicolumn{3}{c}{Gemma~2} && \multicolumn{4}{c}{Gemma~3} \\ 
\cmidrule{2-3}\cmidrule{5-6}\cmidrule{8-10}\cmidrule{12-15}
& Flash & Pro && Flash & Pro && 2B &  9B & 27B  && 1B & 4B & 12B & 27B \\
\midrule
MMLU-Pro & 67.3 & 75.8 && 77.6 & 79.1   && 15.6 & 46.8 & 56.9 && 14.7 & 43.6 & 60.6 & 67.5\\
\midrule
LiveCodeBench & 30.7 & 34.2 && 34.5 & 36.0 && 1.2 & 10.8 & 20.4 && 1.9 & 12.6 & 24.6 & 29.7 \\
Bird-SQL (dev) & 45.6 & 54.4 && 58.7 & 59.3&& 12.2 & 33.8 & 46.7 && 6.4 & 36.3 & 47.9 & 54.4 \\
\midrule
GPQA Diamond & 51.0 & 59.1 && 60.1 & 64.7 && 24.7 & 28.8 & 34.3 && 19.2 & 30.8 & 40.9 & 42.4 \\
\midrule
SimpleQA & 8.6 & 24.9 && 29.9 & 44.3 && 2.8 & 5.3 & 9.2 && 2.2 & 4.0 & 6.3 & 10.0 \\ 
FACTS Grounding & 82.9 & 80.0 && 84.6 & 82.8 && 43.8 & 62.0 & 62.4 && 36.4 & 70.1 & 75.8 & 74.9 \\
 \midrule
 Global MMLU-Lite   & 73.7 & 80.8 && 83.4 & 86.5 && 41.9 & 64.8 & 68.6 && 34.2 & 54.5 & 69.5 &  75.1\\
\midrule
MATH & 77.9 & 86.5 && 90.9 & 91.8 && 27.2 & 49.4 & 55.6 && 48.0 & 75.6 & 83.8 & 89.0 \\
HiddenMath & 47.2 & 52.0 && 63.5 & 65.2 && 1.8 & 10.4 & 14.8 && 15.8 & 43.0 & 54.5 & 60.3 \\
\midrule
MMMU (val)  & 62.3 & 65.9 && 71.7 & 72.7 && - & - & - && - & 48.8 & 59.6 & 64.9 \\
\bottomrule
\end{tabular}
\caption{Performance of instruction fine-tuned (IT) models compared to Gemini 1.5, Gemini 2.0, and Gemma~2 on zero-shot benchmarks across different abilities.}
\label{tab:it_fs}
\end{table*}

\subsection{Standard benchmarks}

In Table~\ref{tab:it_fs}, we show the performance of our final models across a variety of benchmarks compared to our previous model iteration, and Gemini 1.5. 
We do not compare directly with external models that often report their own evaluation settings, since running them in our setting does not guarantee a fair comparison.
We encourage the reader to follow third-party static leaderboards for a fairer comparison across models. 
We include additional evaluations of our models on other benchmarks in the appendix.

\section{Ablations}

In this section, we focus on the impact of our architecture changes, as well as some of the vision abilities new to this model.

\subsection{Pre-training ability probing}

\begin{figure*}[t]
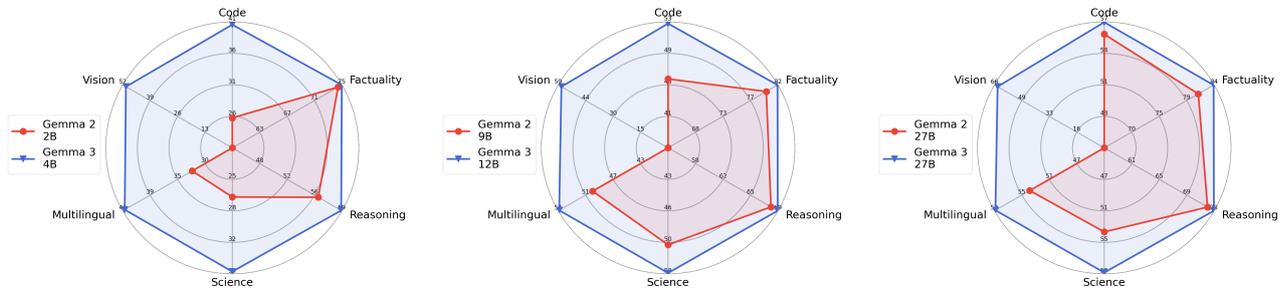

    \centering
    \begin{tabular}{ccc}
       \includegraphics[width=.32\linewidth]{assets/radar_pt_1.pdf} &
       \includegraphics[width=.32\linewidth]{assets/radar_pt_2.pdf} &
       \includegraphics[width=.32\linewidth]{assets/radar_pt_3.pdf}
    \end{tabular}
    \caption{Summary of the performance of different pre-trained models from Gemma~2 and 3 across general abilities. These plots are meant to give a simplified summary and details are in the appendix.}
    \label{fig:pt_summary} 
\end{figure*}

We use several standard benchmarks as probes during pre-training to ensure our models capture general abilities, and in Figure~\ref{fig:pt_summary}, we compare the quality of pre-trained models from Gemma~2 and 3 across these general abilities, namely, science, code, factuality, multilinguality, reasoning, and vision.
The details of the performance across the different public benchmarks used in these plots are summarized in the appendix.
Overall, we see that the new versions improve in most categories, despite the addition of vision.
We particularly focus on multilinguality in this version, and this directly impacts the quality of our models.
However, despite the use of decontamination techniques, there is always a risk of contamination of these probes~\citep{mirzadeh2024gsm}, making more definitive conclusions harder to assess.

\begin{figure}[h]
    \centering
    \includegraphics[width=.8\linewidth]{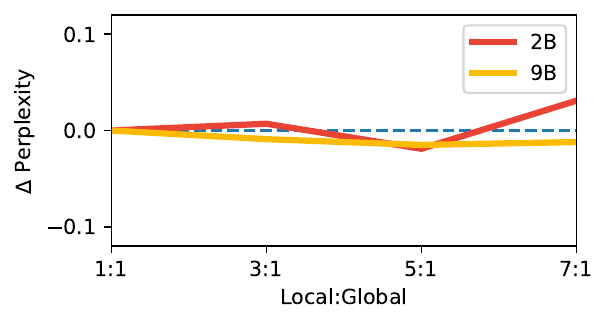}
    \caption{\textbf{Impact of Local:Global ratio} on the perplexity on a validation set. 
    The impact is minimal, even with 7-to-1 local to global. 
    This ablation is run with text-only models.}
    \label{fig:local_global}
\end{figure}

\subsection{Local:Global attention layers}

We measure the impact of changes to local and global self-attention layers on performance and memory consumption during inference.

\noindent\textbf{Local:Global ratio.} 
In Fig.~\ref{fig:local_global}, we compare different ratios of local to global attention layers.
1:1 is used in Gemma~2 models, and 5:1 is  used in Gemma~3.
We observe minimal impact on perplexity when changing this ratio.

\begin{figure}[h]
    \centering
    \includegraphics[width=.8\linewidth]{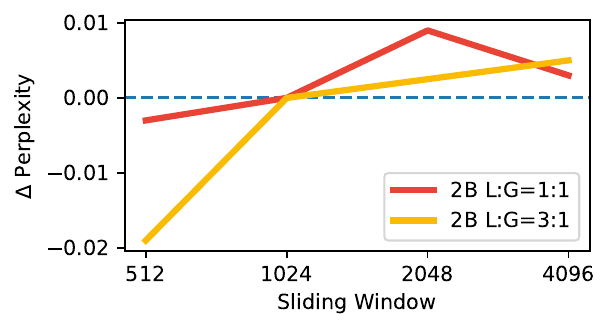}
    \caption{\textbf{Impact of Sliding Window} size on perplexity measured on a validation set. 
    We consider 2 2B models, with 1:1 and 1:3 local to global layer ratios.
    This ablation is run with text-only models.}
    \label{fig:sw}
\end{figure}

\noindent\textbf{Sliding window size.} In Fig.~\ref{fig:sw}, we compare different sliding window sizes for the local attention layers in different global:local ratio configurations.
The sliding window can be reduced significantly without impacting perplexity.

\begin{figure}[h]
    \centering
    \includegraphics[width=\linewidth]{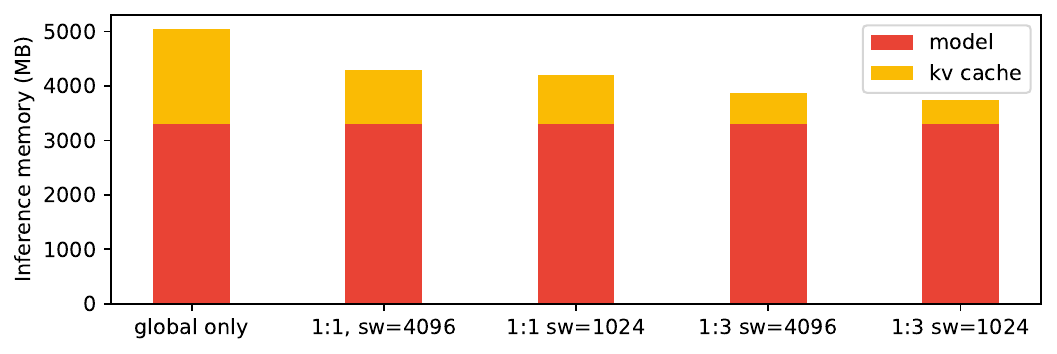}
    \caption{\textbf{Model versus KV cache memory} during inference with a pre-fill KV cache of size 32k. 
    We consider a 2B model with different local to global ratios and sliding window sizes (sw).
    We compare to global only, which is the standard used in Gemma 1 and Llama.
    This ablation is run with a text-only model.}
    \label{fig:kv}
\end{figure}

\begin{figure}[h]
    \centering
    \includegraphics[width=.8\linewidth]{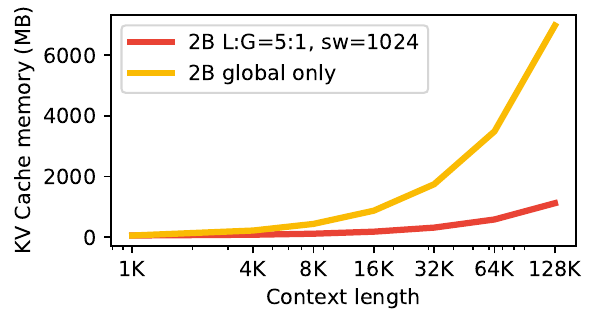}
    \caption{\textbf{KV cache memory versus context length.} We show the memory usage of the KV cache for our architecture (L:G=5:1, sw=1024) and a transformer with global attention only -- as used in LLaMa or Gemma 1.}
    \label{fig:kv_lc}
\end{figure}

\noindent\textbf{Impact on KV cache memory.}
In Fig.~\ref{fig:kv}, we show the balance between the memory used by the model and the KV cache during inference with a context of 32k tokens. 
The ``global only'' configuration is the standard configuration used across most dense models.
The ``1:1, sw=4096'' is used in Gemma~2.
We observe that the ``global only'' configuration results in a memory overhead of 60\%, while this is reduced to less than 15\% with 1:3 and sliding windows of 1024 (``sw=1024'').
In Fig.~\ref{fig:kv_lc}, we compute the memory used by the KV cache as a function of the context length with either our 2B architecture (L:G=5:1, sw=1024) versus a ``global only'' 2B model.

\subsection{Enabling long context}

\begin{figure}[h]
    \centering
    \includegraphics[width=\linewidth]{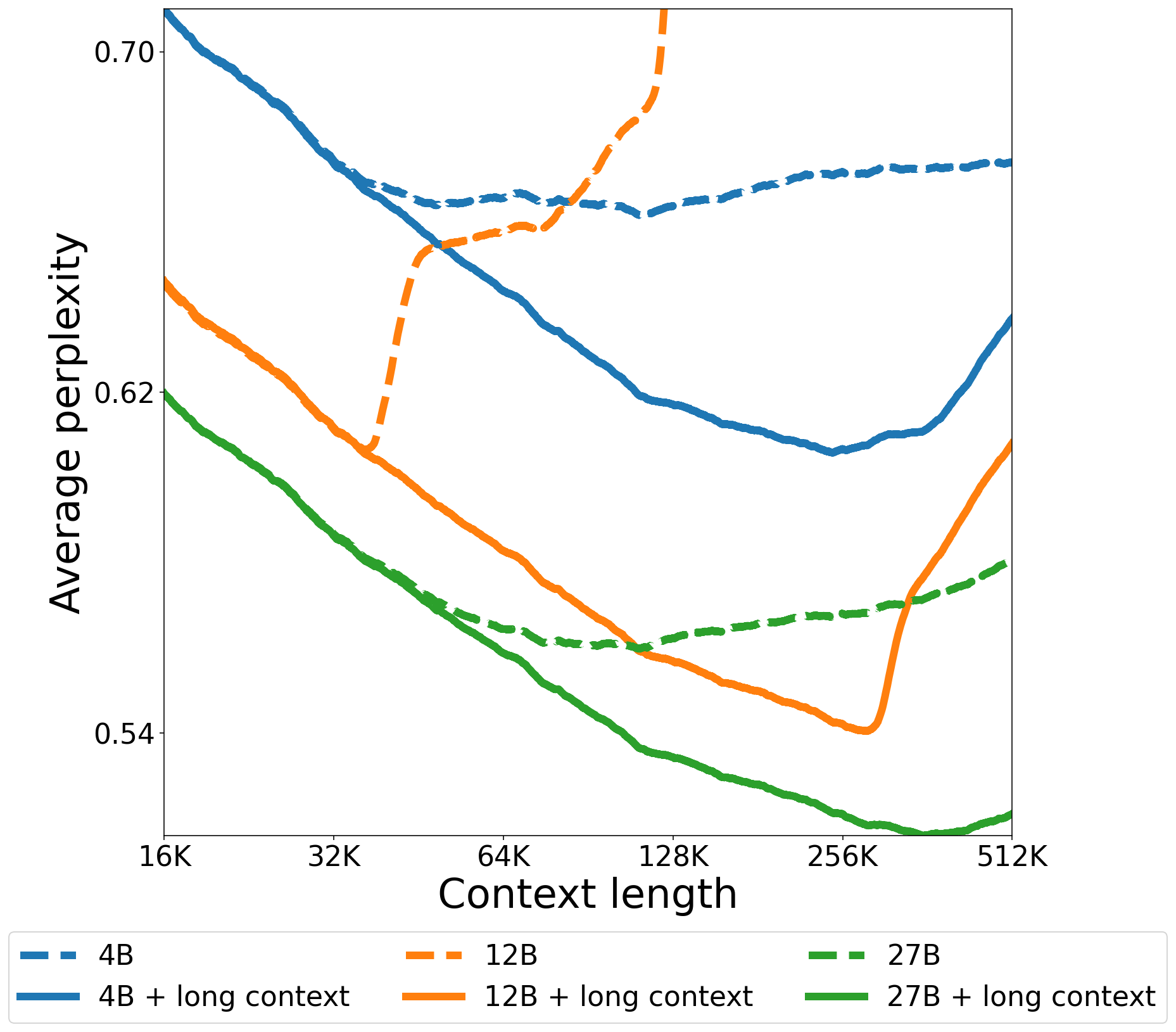}
    \caption{\textbf{Long context} performance of pre-trained models before and after RoPE rescaling.}
    \label{fig:lc_all_sizes}
\end{figure}

Instead of training with 128K sequences from scratch, we pre-train our models with 32K sequences and then scale the 4B, 12B, and 27B models up to 128K tokens at the end of pre-training while rescaling RoPE \citep{chen2023extending}. We find a scaling factor of 8 to work well in practice. Note that compared to Gemma 2, we have also increased the RoPE base frequency of global self-attention layers from 10k to 1M, while keeping 10k for the local self-attention layers.
In Figure~\ref{fig:lc_all_sizes}, we show the impact on perplexity for different context lengths. Our models generalize to 128K, but rapidly degrade as we continue to scale.

\subsection{Small versus large teacher} 

\begin{figure}[h]
    \centering
    \includegraphics[width=.7\linewidth]{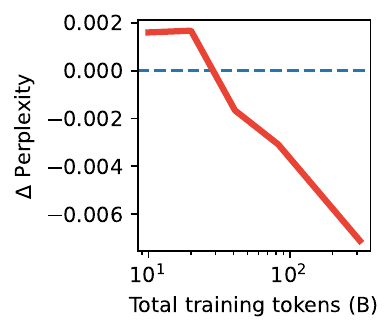}
    \caption{\textbf{Small versus large teacher.} Relative difference of perplexity when using a small and large teacher as a function of the token size of training. Smaller numbers means distilling from a larger teacher is better.}
    \label{fig:teachers}
\end{figure}

A common finding is that, to train a small model, it is preferable to distill from a smaller teacher.
We suspect this is because these studies are often performed in settings where the regularization effect of using a worse teacher surpasses the benefit of using a better teacher.
We train a student with 2 teachers of different sizes, one large and one small, for different training horizons. In Fig.~\ref{fig:teachers}, we observe that for short training horizons, the smaller teacher is better, but the trend is reversed for longer training.

\subsection{Vision encoder}

\begin{table}[h]
\centering
\setlength{\tabcolsep}{.45em}
\begin{tabular}{@{}c  ccc @{}}  
\toprule
Resolution & DocVQA & InfoVQA & TextVQA \\
\midrule
256 & 31.9 & 23.1 & 44.1 \\
448  & 45.4  & 31.6 & 53.5 \\
896  & 59.8 & 33.7 & 58.0 \\
\bottomrule
\end{tabular}
\caption{
\textbf{Impact of image encoder input resolution.}
We measure performance using a short schedule 2B Gemma model on a few evaluation benchmarks to observe the effect of input image resolution on vision encoder pre-training. 
}
\label{tab:resolution}
\end{table}

\noindent\textbf{Impact of image resolution.} 
We use a vision encoder based on SigLIP ~\citep{zhai2023sigmoid}. 
The vision encoder is frozen, and only the language model is trained. 
Each image in this multimodal data is represented by 256 image tokens from the respective vision encoder. 
The higher resolution encoders thus use average pooling to reduce their output to 256 tokens. 
For instance, the 896 resolution encoder has a 4x4 average pooling on its output.
As shown in Table \ref{tab:resolution}, higher resolution encoders perform better than smaller ones.

\begin{table}[h]
\centering
\setlength{\tabcolsep}{.45em}
\begin{tabular}{@{}c ccc @{}}  
\toprule
& DocVQA & InfoVQA & TextVQA \\
\midrule 
4B  & 72.8 & 44.1 & 58.9 \\
4B w/ P\&S    & 81.0 & 57.0 & 60.8 \\
\multicolumn{1}{c}{$\Delta$} & \footnotesize\color{blue!70!black} (+8.2) & \footnotesize\color{blue!70!black} (+12.9) & \footnotesize\color{blue!70!black} (+1.9) \\
\midrule 
27B & 85.6 & 59.4 & 68.6 \\
27B w/ P\&S    & 90.4 & 76.4 & 70.2 \\
\multicolumn{1}{c}{$\Delta$} & \footnotesize\color{blue!70!black} (+4.8) & \footnotesize\color{blue!70!black} (+17.0) & \footnotesize\color{blue!70!black} (+1.6) \\
\bottomrule
\end{tabular}
\caption{
\textbf{Impact of P\&S.}
4-shot evaluation results on the valid set, with and without P\&S on a pre-trained checkpoint. Boosts are on tasks associated with images with varying aspect ratios, or involving reading text on images.
}
\label{tab:AIW}
\end{table}

\noindent\textbf{Pan \& Scan.} 
P\&S enables capturing images at close to their native aspect ratio and image resolution. In Table~\ref{tab:AIW}, we compare our 27B IT model with and without P\&S. 
As expected, the ability to treat images with close to native resolution greatly helps with tasks that require some form of reading text on images, which is particularly important for visual language models.

\section{Memorization and Privacy} \label{sec:memorization}

Large language models may produce near-copies of some text used in training~\citep{carlini2021extracting,carlini2022quantifying,ippolito2022preventing,biderman2023emergent,nasr2023scalable}. Several prior reports have released audits that quantify this risk by measuring the memorization rate~\citep{geminiteam2023gemini,geminiteam2024gemini,gemmateam2024gemma,gemmateam2024gemma2,anil2023palm,chowdhery2022palm,dubey2024llama}. This ``memorization rate''\footnote{"We do not state or imply [here] that a model "contains" its training data in the sense that there is a copy of that data in the model.  Rather, a model memorizes attributes of its training data such that in certain cases it is statistically able to generate such training data when following rules and using information about features of its training data that it does contain."} is defined as the ratio of generations from the model that match its training data compared to all model generations using the following setup. We follow the methodology described in~\citet{gemmateam2024gemma2} to measure it. Specifically, we subsample a large portion of training data distributed uniformly across different corpora and test for discoverable extraction~\citep{nasr2023scalable} of this content using a prefix of length 50 and a suffix of length 50. We denote text as either ``exactly memorized'' if all tokens in the continuation match the source suffix or ``approximately memorized'' if they match up to an edit distance of 10\%.

\begin{figure}
    \centering
    \includegraphics[width=\linewidth]{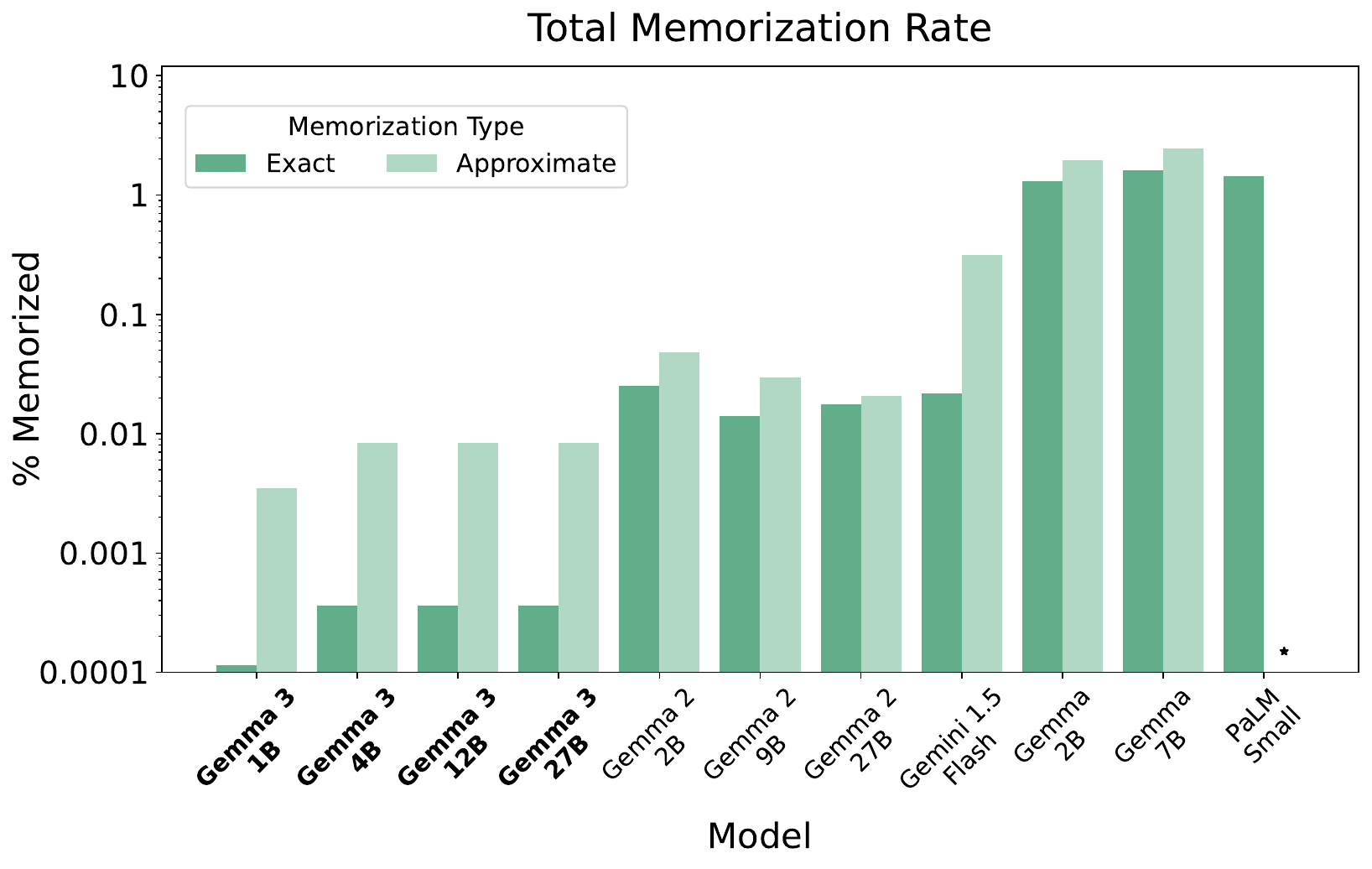}
    \caption{Total memorization rates for both exact and approximate memorization. Gemma 3 models memorize significantly less than all prior models. *No results for approximate memorization on these models.}
    \label{fig:memorization}
\end{figure}

Figure~\ref{fig:memorization} compares the memorization rates across Gemma and Gemini models; these models are ordered in reverse chronological order, with the newest Gemma 3 models on the left. We find that Gemma 3 models memorize long-form text at a much lower rate than prior models (note the log y-axis). We observe only a marginal difference in the memorization rates between the 4B, 12B, and 27B models, with 1B memorizing less than these larger models. Further, we find that a larger proportion of text is characterized as approximately memorized, with a relative increase in approximate memorization compared to exact memorization of roughly 24x on average.

We also study the rate at which the generations may contain personal information. To identify potentially personal information, we use the Google Cloud Sensitive Data Protection (SDP) service.\footnote{https://cloud.google.com/sensitive-data-protection} SDP uses broad detection rules to identify text that may contain personal information. SDP is designed to have high recall and does not consider the context in which the information may appear, which leads to many false positives. Thus, we are likely overestimating the true amount of potentially personal information contained in the outputs classified as memorized. SDP also provides broad severity levels: low, medium, and high. We classify text as personal if SDP classifies it as personal information at any severity level. We observed no personal information in the outputs characterized as memorization for all Gemma 3 models. This indicates a low rate of personal data, below our detection thresholds, in outputs classified as memorization.

\section{Responsibility, Safety, Security}
\label{safety}

Responsibility, safety, and security are of utmost importance in the development of Gemma models. To reduce risks to Gemma 3 users, we have continued to integrate enhanced internal safety processes that span the development workflow, in line with recent Google AI models~\citep{geminiteam2024gemini}. This focuses on safety mitigation at training time, and robust and transparent model evaluations for the new image-to-text capabilities we have introduced.


\subsection{Governance \& Assessment}

Our approach to assessing the benefits and risks of Gemma is reflective of that outlined for Gemma 1~\citep{gemmateam2024gemma}, taking into account the changes in supported modalities. We continue to believe that openness in AI can spread the benefits of these technologies across society, but must be evaluated against the risk of malicious uses that can cause harm on both individual and institutional levels~\citep{weidinger2021ethicalsocialrisksharm}. Since the inaugural Gemma launch, we have seen these models drive a number of socially beneficial applications, such as our own ShieldGemma 2, a 4B image safety classifier built with Gemma 3, which provides a ready-made solution for image safety, outputting safety labels across dangerous content, sexually explicit, and violence categories.

Releasing Gemma 3 models required specific attention to changes in model capabilities and close monitoring of the evolving risks of existing multimodal LLMs~\citep{lin2024mallademystifyingrealworldlarge}, as well as an understanding of the ways in which models are being used in the wild. Although we are yet to receive any reports of malicious use for Gemma, we remain committed to investigating any such reporting, and work with the academic and developer communities, as well as conduct our own monitoring, to flag such cases.

Despite advancements in capabilities, we believe that, given the number of larger powerful open models available, this release will have a negligible effect on the overall risk landscape.

\subsection{Safety policies and train-time mitigations}

A key pillar of Gemma’s approach to safety is to align fine-tuned models with Google’s safety policies, in line with Gemini models~\citep{geminiteam2023gemini}. They are designed to help prevent our models from generating harmful content, i.e.,
\begin{itemize}
    \item 
    Child sexual abuse and exploitation
    \item
    Revealing personally identifiable information that can lead to harm (e.g., Social Security numbers)
    \item
    Hate speech and harassment
    \item
    Dangerous or malicious content (including promoting self-harm or instructing in harmful activities)
    \item
    Sexually explicit content
    \item
    Medical advice that runs contrary to scientific or medical consensus
\end{itemize}
We undertook considerable safety filtering of our pre-training data to reduce the likelihood of our pre-trained and fine-tuned checkpoints producing harmful content. For fine-tuned models, we also use both SFT and RLHF to steer the model away from undesirable behavior.



\subsection{Assurance Evaluations}

We also run our IT models through a set of baseline assurance evaluations to understand the potential harms that our models can cause. As we champion open models, we also recognize that the irreversible nature of weight releases requires rigorous risk assessment. Our internal safety processes are designed accordingly, and for previous Gemma models we have also undertaken evaluations of capabilities relevant to extreme risks~\citep{shevlane2023modelevaluationextremerisks, phuong2024evaluatingfrontiermodelsdangerous}.  As we continue to develop and share open models, we will follow the heuristic that thoroughly evaluating a more capable model often provides sufficient assurance for less capable ones. As such, we prioritised a streamlined set of evaluations for Gemma 3, reserving in-depth dangerous capability assessments for cases where a specific model may present a potentially heightened risk (as described below on CBRN evaluations). We balance development speed with targeted safety testing, ensuring our evaluations are well-focused and efficient, while upholding the commitments laid out in our Frontier Safety Framework.

\subsubsection*{Baseline Evaluations}

Baseline assurance captures the model violation rate for safety policies, using a large number of synthetic adversarial user queries, and human raters to label the answers as policy violating or not. Overall, Gemma 3 violation rate is significantly low overall on these safety policies.

\subsubsection*{Chemical, Biological, Radiological and Nuclear (CBRN) knowledge }

Owing to enhanced performance on STEM-related tasks, we evaluated knowledge relevant to biological, radiological, and nuclear risks using an internal dataset of closed-ended, knowledge-based multiple choice questions. For evaluations of chemical knowledge, we employed a closed-ended knowledge-based approach on chemical hazards developed by \cite{MAG} Our evaluation suggests that the knowledge of Gemma 3 models in these domains is low.

\subsection{Our approach to responsible open models}

Designing safe, secure, and responsible applications requires a system-level approach, working to mitigate risks associated with each specific use case and environment. We will continue to adopt assessments and safety mitigations proportionate to the potential risks from our models, and will only share these with the community when we are confident that the benefits significantly outweigh the foreseeable risks.

\section{Discussion and Conclusion}

In this work, we have presented Gemma~3, the latest addition to the Gemma family of open language models for text, image, and code. 
In this version, we focus on adding image understanding and long context while improving multilinguality and STEM-related abilities. 
Our model sizes and architectures are designed to be compatible with standard hardware, and most of our architecture improvements are tailored to fit this hardware while maintaining performance.

\bibliography{main}

\noindent\textbf{Core contributors} \\
Aishwarya Kamath$^*$\blfootnote{$^*$ co-first authors.} \\
Johan Ferret$^*$ \\
Shreya Pathak$^*$ \\
Nino Vieillard$^*$ \\
Ramona Merhej$^*$ \\
Sarah Perrin$^*$ \\
Tatiana Matejovicova$^*$ \\
Alexandre Ramé$^*$ \\
Morgane Rivière$^*$ \\
Louis Rouillard$^*$ \\
Thomas Mesnard$^*$ \\
Geoffrey Cideron$^*$ \\
Jean-bastien Grill$^*$ \\
Sabela Ramos$^*$ \\
Edouard Yvinec$^*$ \\
Michelle Casbon$^*$ \\
Etienne Pot \\
Ivo Penchev \\
Gaël Liu \\
Francesco Visin \\
Kathleen Kenealy \\
Lucas Beyer \\
Xiaohai Zhai \\
Anton Tsitsulin \\
Robert Busa-Fekete \\
Alex Feng \\
Noveen Sachdeva \\
Benjamin Coleman \\
Yi Gao \\
Basil Mustafa \\
Iain Barr \\
Emilio Parisotto \\
David Tian \\
Matan Eyal \\
Colin Cherry \\
Jan-Thorsten Peter \\
Danila Sinopalnikov \\
Surya Bhupatiraju \\
Rishabh Agarwal \\
Mehran Kazemi \\
Dan Malkin \\
Ravin Kumar \\
David Vilar \\
Idan Brusilovsky \\
Jiaming Luo \\
Andreas Steiner \\

\noindent\textbf{Contributors (alphabetical order)} \\
Abe Friesen \\
Abhanshu Sharma \\
Abheesht Sharma \\
Adi Mayrav Gilady \\
Adrian Goedeckemeyer \\
Alaa Saade \\
Alex Feng \\
Alexander Kolesnikov \\
Alexei Bendebury \\
Alvin Abdagic \\
Amit Vadi \\
András György \\
André Susano Pinto \\
Anil Das \\
Ankur Bapna \\
Antoine Miech \\
Antoine Yang \\
Antonia Paterson \\
Ashish Shenoy \\
Ayan Chakrabarti \\
Bilal Piot \\
Bo Wu \\
Bobak Shahriari \\
Bryce Petrini \\
Charlie Chen \\
Charline Le Lan \\
Christopher A. Choquette-Choo \\
CJ Carey \\
Cormac Brick \\
Daniel Deutsch \\
Danielle Eisenbud \\
Dee Cattle \\
Derek Cheng \\
Dimitris Paparas \\
Divyashree Shivakumar Sreepathihalli \\
Doug Reid \\
Dustin Tran \\
Dustin Zelle \\
Eric Noland \\
Erwin Huizenga \\
Eugene Kharitonov \\
Frederick Liu \\
Gagik Amirkhanyan \\
Glenn Cameron \\
Hadi Hashemi \\
Hanna Klimczak-Plucińska \\
Harman Singh \\
Harsh Mehta \\
Harshal Tushar Lehri \\
Hussein Hazimeh \\
Ian Ballantyne \\
Idan Szpektor \\
Ivan Nardini \\
Jean Pouget-Abadie \\
Jetha Chan \\
Joe Stanton \\
John Wieting \\
Jonathan Lai \\
Jordi Orbay \\
Joseph Fernandez \\
Josh Newlan \\
Ju-yeong Ji \\
Jyotinder Singh \\
Kat Black \\
Kathy Yu \\
Kevin Hui \\
Kiran Vodrahalli \\
Klaus Greff \\
Linhai Qiu \\
Marcella Valentine \\
Marina Coelho \\
Marvin Ritter \\
Matt Hoffman \\
Matthew Watson \\
Mayank Chaturvedi \\
Michael Moynihan \\
Min Ma \\
Nabila Babar \\
Natasha Noy \\
Nathan Byrd \\
Nick Roy \\
Nikola Momchev \\
Nilay Chauhan \\
Noveen Sachdeva \\
Oskar Bunyan \\
Pankil Botarda \\
Paul Caron \\
Paul Kishan Rubenstein \\
Phil Culliton \\
Philipp Schmid \\
Pier Giuseppe Sessa \\
Pingmei Xu \\
Piotr Stanczyk \\
Pouya Tafti \\
Rakesh Shivanna \\
Renjie Wu \\
Renke Pan \\
Reza Rokni \\
Rob Willoughby \\
Rohith Vallu \\
Ryan Mullins \\
Sammy Jerome \\
Sara Smoot \\
Sertan Girgin \\
Shariq Iqbal \\
Shashir Reddy \\
Shruti Sheth \\
Siim Põder \\
Sijal Bhatnagar \\
Sindhu Raghuram Panyam \\
Sivan Eiger \\
Susan Zhang \\
Tianqi Liu \\
Trevor Yacovone \\
Tyler Liechty \\
Uday Kalra \\
Utku Evci \\
Vedant Misra \\
Vincent Roseberry \\
Vlad Feinberg \\
Vlad Kolesnikov \\
Woohyun Han \\
Woosuk Kwon \\
Xi Chen \\
Yinlam Chow \\
Yuvein Zhu \\
Zichuan Wei \\
Zoltan Egyed \\

\noindent\textbf{Support} \\
Victor Cotruta \\
Minh Giang \\
Phoebe Kirk \\
Anand Rao \\
Kat Black \\
Nabila Babar \\
Jessica Lo \\
Erica Moreira \\
Luiz Gustavo Martins \\
Omar Sanseviero \\
Lucas Gonzalez \\
Zach Gleicher \\
Tris Warkentin \\

\noindent\textbf{Sponsors} \\
Vahab Mirrokni \\
Evan Senter \\
Eli Collins \\
Joelle Barral \\
Zoubin Ghahramani \\
Raia Hadsell \\
Yossi Matias \\
D. Sculley \\
Slav Petrov \\
Noah Fiedel \\
Noam Shazeer \\
Oriol Vinyals \\
Jeff Dean \\
Demis Hassabis \\
Koray Kavukcuoglu \\
Clement Farabet \\

\noindent\textbf{Technical advisors} \\
Elena Buchatskaya \\
Jean-Baptiste Alayrac \\
Rohan Anil \\
Dmitry (Dima) Lepikhin \\
Sebastian Borgeaud \\
Olivier Bachem \\

\noindent\textbf{Lead} \\
Armand Joulin \\

\noindent\textbf{Technical leads} \\
Alek Andreev \\
Cassidy Hardin \\
Robert Dadashi \\
Léonard Hussenot \\

\section*{Appendix}

\subsection*{Details of pre-trained performances.}

\begin{table}[h]
\centering
\setlength{\tabcolsep}{.25em}
\begin{tabular}{@{}l c ccc  c cccc@{}}  
\toprule
& \multicolumn{3}{c}{Gemma~2} && \multicolumn{4}{c}{Gemma 3} \\ 
\cmidrule{2-4}\cmidrule{6-9}
&  2B &  9B & 27B  && 1B & 4B & 12B & 27B \\
\midrule
HellaS   & 72.9 & 81.9 & \textbf{86.4} && 62.3 & 77.2 & 84.2 & 85.6\\
BoolQ    & 75.6 & 77.5 & 76.2 && 63.2 & 72.3 & 78.8 & \textbf{82.4}\\
\midrule
PIQA     & 78.1 & 81.9 & \textbf{83.5} && 73.8 & 79.6 & 81.8 & 83.3\\
SIQA     & 51.8 & 53.3 & 53.8 && 48.9 & 51.9 & 53.4 & \textbf{54.9}\\
TQA      & 60.2 & 76.5 & 83.8 && 39.8 & 65.8 & 78.2 & \textbf{85.5}\\
NQ       & 17.2 & 29.2 & 34.7 && 9.48 & 20.0 & 31.4 & \textbf{36.1}\\
\midrule
ARC-C    & 55.8 & 69.1 & \textbf{71.4} && 38.4 & 56.2 & 68.9 & 70.6\\
ARC-E    & 80.6 & 88.3 & 88.6 && 73.0 & 82.4 & 88.3 & \textbf{89.0}\\
WinoG    & 65.4 & 73.9 & \textbf{79.4} && 58.2 & 64.7 & 74.3 & 78.8\\
BBH      & 42.4 & 69.4 & 74.8 && 28.4 & 50.9 & 72.6 & \textbf{77.7}\\
Drop     & 53.2 & 71.5 & 75.2 && 42.4 & 60.1 & 72.2 & \textbf{77.2}\\
\bottomrule
\end{tabular}
\caption{
Factuality, common-sense performance and reasoning after pre-training phase.
}
\label{tab:fcs}
\end{table}

\noindent\textbf{Factuality and common-sense.} In Table~\ref{tab:fcs}, we report the performance of our new pre-trained benchmarks compared to previous versions.
We consider several standard benchmarks, namely HellaSwag~\citep{zellers-etal-2019-hellaswag}, BoolQ~\citep{boolq}, PIQA~\citep{piqa}, SIQA~\citep{siqa}, TriviaQA~\citep{triviaqa}, Natural Questions~\citep{natural-questions}, ARC-C and ARC-E~\citep{chollet2019measure}, WinoGrande~\citep{winogrande}, BBH~\citep{bbhard}, DROP~\citep{dua-etal-2019-drop}. Evaluation details are described in Table~\ref{tab:eval_detail_text}.
Overall, our models are in the same ballpark as Gemma~2, which is encouraging since these abilities are not the focus of the improvements brought in this version. 

\begin{table}[h]
\centering
\setlength{\tabcolsep}{.25em}
\begin{tabular}{@{}l  ccc c  ccc@{}}  
\toprule
& \multicolumn{3}{c}{Gemma~2} && \multicolumn{3}{c}{Gemma 3} \\ 
\cmidrule{2-4}\cmidrule{6-8}
&  2B &  9B & 27B  && 4B & 12B & 27B \\
\midrule
MMLU    & 52.2 & 71.2 & 75.2 && 59.6 & 74.5 & \textbf{78.6}\\
MMLUpro & 22.2 & 43.7 & 49.4 && 29.2 & 45.3 & \textbf{52.2}\\
AGIE    & 31.6 & 53.1 & 55.1 && 42.1 & 57.4 & \textbf{66.2}\\
MATH    & 16.4 & 36.4 & 42.1 && 24.2 & 43.3 & \textbf{50.0}\\
GSM8K   & 25.0 & 70.2 & 74.6 && 38.4 & 71.0 & \textbf{82.6}\\
GPQA Diamond   & 12.5 & 24.8 & \textbf{26.3} && 15.0 & 25.4 & 24.3\\
MBPP    & 31.0 & 51.2 & 60.8 && 46.0 & 60.4 & \textbf{65.6}\\
HumanE  & 19.5 & 40.2 & \textbf{51.2} && 36.0 & 45.7 & 48.8\\
\bottomrule
\end{tabular}
\caption{
STEM and code performance after pre-training phase.
}
\label{tab:stem}
\end{table}

\noindent\textbf{STEM and code.} The details of our performance on STEM and Code are in Table~\ref{tab:stem}. We consider several standard benchmarks, namely MMLU~\citep{mmlu}, MMLU-Pro~\citep{wang2024mmlu}, AGIEval~\citep{zhong2023agieval}, MATH~\citep{hendrycksmath2021}, GSM8K~\citep{gsm8k}, GPQA~\citep{Rein2023GPQAAG}, MBPP~\citep{mbpp}, HumanEval~\citep{humaneval}. Evaluation details are described in Table~\ref{tab:eval_detail_text}.
Overall we see a consistent improvement over STEM abilities across our pre-trained models. On code, we see a similar improvement for the 4B and 12B models but not on the 27B.

\begin{table}[h]
\centering
\setlength{\tabcolsep}{.6em}
\begin{tabular}{@{}l ccc@{}}  
\toprule
& 4B & 12B & 27B \\
\midrule
COCO caption       & 102  & 111  & \textbf{116} \\
DocVQA             & 72.8 & 82.3 & \textbf{85.6}\\
InfoVQA            & 44.1 & 54.8 & \textbf{59.4}\\
MMMU               & 39.2 & 50.3 & \textbf{56.1}\\
TextVQA            & 58.9 & 66.5 & \textbf{68.6}\\
RealWorldQA        & 45.5 & 52.2 & \textbf{53.9}\\
ReMI               & 27.3 & 38.5 & \textbf{44.8}\\
AI2D               & 63.2 & 75.2 & \textbf{79.0}\\
ChartQA            & 63.6 & 74.7 & \textbf{76.3}\\
VQAv2              & 63.9 & 71.2 & \textbf{72.9}\\
BLINK              & 38.0 & 35.9 & \textbf{39.6}\\
OK-VQA             & 51.0 & 58.7 & \textbf{60.2}\\
TallyQA            & 42.5 & 51.8 & \textbf{54.3}\\
SpatialSense VQA   & 50.9 & \textbf{60.0} & 59.4\\
CountBench VQA      & 26.1 & 17.8 & \textbf{68.0}\\
\bottomrule
\end{tabular}
\caption{
Multimodal performance after pre-training phase. The scores are on the val split of each dataset without P\&S.
}
\label{tab:multimodal}
\end{table}

\noindent\textbf{Image understanding.} In Table~\ref{tab:multimodal}, we report performance across a variety of visual question answer benchmarks for the different models that were trained with a vision encoder, namely COCO Caption~\citep{Chen2015MicrosoftCC}, DocVQA~\citep{Mathew2020DocVQAAD}, InfographicVQA~\citep{mathew2022infographicvqa}, MMMU~\citep{Yue2023MMMUAM}, TextVQA~\citep{singh2019towards}, RealWorldQA~\citep{RealWorldQA}, ReMI~\citep{Kazemi2024ReMIAD}, AI2D~\citep{Kembhavi2016ADI}, ChartQA~\citep{masry-etal-2022-chartqa}, VQA v2~\citep{balanced_vqa_v2}, BLINK~\citep{Fu2024BLINKML}, OK-VQA~\citep{okvqa}, TallyQA~\citep{Acharya2018TallyQAAC}, SpatialSense VQA~\citep{Yang2019SpatialSenseAA}, CountBench VQA~\citep{Paiss2023TeachingCT}.  Evaluation details are described in Table~\ref{tab:eval_detail_mm}. 

\begin{table}[h]
\centering
\small
\setlength{\tabcolsep}{.25em}
\begin{tabular}{@{}l  ccc c  ccc@{}}  
\toprule
& \multicolumn{3}{c}{PaliGemma 2}       && \multicolumn{3}{c}{Gemma 3} \\ 
\cmidrule{2-4}\cmidrule{6-8}
                &  2B &  9B & 27B       && 4B & 12B & 27B \\
\midrule
DocVQA          & 81.6 & 86.3 & 85.1    && 86.1 & 89.0 & \textbf{89.5}\\
InfoVQA         & 41.4 & 53.1 & 50.2    && 55.6 & 61.6 & \textbf{64.6}\\
TextVQA         & 76.3 & 76.3 & 75.1    && 79.1 & 81.6 & \textbf{83.2}\\
ChartQA         & 70.7 & 79.1 & 71.3    && 79.8 & 83.5 & 83.4\\
\midrule
AI2D            & 76.0 & 84.4 & 84.6    && 80.9 & 85.6 & \textbf{86.5}\\
OKVQA           & 64.1 & 68.6 & 70.6    && 65.2 & 69.3 & \textbf{71.1}\\
CountBenchQA    & 82.0 & 85.3 & 87.4    && 79.4 & 83.5 & \textbf{87.8}\\
\midrule
COCO caption    & 143. & \textbf{145.} & \textbf{145.}    && 143. & 143. & 144.\\
VQAv2           & 84.8 & \textbf{85.8} & \textbf{85.8}    && 84.1 & 84.9 & 85.1\\
Tally QA        & 80.6 & \textbf{82.4} & 82.1    && 79.0 & 81.3 & 81.7\\
\bottomrule
\end{tabular}
\caption{
Performance of pre-trained checkpoints after fine-tuning on multi-modal benchmarks (without P\&S). PaliGemma~2 was transferred at 896x896 resolution for the first four benchmarks, and at 448x448 resolution for the others.
}
\label{tab:pg2}
\end{table}

\noindent\textbf{Comparison to PaliGemma 2.}
We fine-tune multimodal Gemma 3 pre-trained checkpoints following the protocol from~\cite{steiner2024paligemma2} -- only learning rate is swept, otherwise the same transfer settings are used. The results in Table~\ref{tab:pg2} show that Gemma~3 excels at benchmarks involving document understanding, even outperforming the \emph{larger} PaliGemma~2 variant. Note that due to average pooling in the vision encoder the Gemma~3 4B and 12B models are about 10x cheaper to transfer compared with the PaliGemma~2 9B and 27B models at the same 896 x 896 resolution. Gemma~3 also performs better on AI2D and OKVQA, but PaliGemma~2 performs slightly better on VQAv2 and COCO caption.

\begin{table}[h]
\centering
\setlength{\tabcolsep}{.22em}
\begin{tabular}{@{}l  ccc c  cccc@{}}  
\toprule
& \multicolumn{3}{c}{Gemma~2} && \multicolumn{4}{c}{Gemma 3} \\ 
\cmidrule{2-4}\cmidrule{6-9}
&  2B &  9B & 27B  && 1B & 4B & 12B & 27B \\
\midrule
MGSM          & 18.7 & 57.3 & 68.0 && 2.04 & 34.7 & 64.3 & \textbf{74.3}\\
GMMLU         & 43.3 & 64.0 & 69.4 && 24.9 & 57.0 & 69.4 & \textbf{75.7}\\
\wmtpp        & 38.8 & 50.3 & 53.0 && 36.7 & 48.4 & 53.9 & \textbf{55.7}\\
Flores        & 30.2 & 41.3 & 44.3 && 29.5 & 39.2 & 46.0 & \textbf{48.8}\\
XQuAD         & 53.7 & 72.2 & 73.9 && 43.9 & 68.0 & 74.5 & \textbf{76.8}\\
ECLeKTic      & 8.29  & 14.0  & 17.1  && 4.69 & 11.0 & 17.2 & \textbf{24.4}\\
IndicGB       & 47.4 & 59.3 & 62.1 && 41.4 & 57.2 & 61.7 & \textbf{63.4}\\
\bottomrule 
\end{tabular} 
\caption{
Multilingual performance after the pre-training phase. IndicGenBench is an average over benchmarks reported in Table~\ref{tab:multilingual-indic-pt}.
}
\label{tab:multilingual-PT} 
\end{table}
    
\noindent\textbf{Multilinguality.}
In Table~\ref{tab:multilingual-PT} we report the performance of the pre-trained models on multilingual tasks.
We apply in-context learning with multi-shot prompting and present results on the following benchmarks: MGSM~\citep{shi2023language}, Global-MMLU-Lite~\citep{singh2024globalmmluunderstandingaddressing}, WMT24++~\citep{deutsch2025wmt24expandinglanguagecoverage}, FLoRes~\citep{goyal2022flores}, XQuAD~\citep{artetxe-etal-2020-cross}, ECLeKTic~\citep{goldman2025eclekticnovelchallengeset}, IndicGenBench~\citep{singh2024indicgenbench}, XOR QA~\citep{asai2020xor}. Evaluation details are described in Table~\ref{tab:eval_detail_text}.

\begin{table}[h]
\centering
\setlength{\tabcolsep}{.22em}
\begin{tabular}{@{}l  ccc c  cccc@{}}  
\toprule
& \multicolumn{3}{c}{Gemma~2} && \multicolumn{4}{c}{Gemma 3} \\ 
\cmidrule{2-4}\cmidrule{6-9}
&  2B &  9B & 27B  && 1B & 4B & 12B & 27B \\
\midrule
XQuAD Indic  		& 54.3 & 73.1 & 74.9 && 43.1 & 68.3 & 75.2 & \textbf{77.8}\\
XORQA in-en 		& 66.2 & 69.3 & \textbf{72.5} && 56.3 & 68.3 & 69.8 & 70.4\\
XORQA in-xx 		& 31.2 & 40.8 & 44.3 && 27.1 & 39.8 & 43.8 & \textbf{46.0}\\
Flores Indic 		& 38.1 & 54.0 & 56.9 && 39.0 & 52.3 & 58.0 & \textbf{59.5}\\
\bottomrule 
\end{tabular} 
\caption{
Detailed IndicGenBench performance after the pre-training phase.
}
\label{tab:multilingual-indic-pt} 
\end{table}

\noindent\textbf{Long context.} In Table~\ref{tab:pt_fs2} we report the performance of pre-trained and fine-tuned models on long context benchmarks. We include RULER~\citep{hsieh2024ruler} and MRCR~\citep{vodrahalli2024michelangelo} benchmarks evaluating at 32K and 128K sequence lengths.

\begin{table}[h!]
\centering
\setlength{\tabcolsep}{.2em}
\begin{tabular}{@{}l c ccc c ccc @{}}
\toprule
& & \multicolumn{3}{c}{Gemma 3 PT} && \multicolumn{3}{c}{Gemma 3 IT} \\ 
\cmidrule{3-5}\cmidrule{7-9}
& Context & 4B & 12B & 27B && 4B & 12B & 27B \\
\midrule
RULER & 32K  & 67.1	& \textbf{90.6} & 85.9 && 61.4 & 80.3 & \textbf{91.1}\\
RULER & 128K & 51.7	& \textbf{80.7} & 72.9 && 46.8 & 57.1 & \textbf{66.0}\\
\midrule
MRCR & 32K  & 44.7	& 59.8 & \textbf{63.2} && 49.8 & 53.7 & \textbf{63.2}\\
MRCR & 128K & 40.6	& 56.9 & \textbf{60.0} && 44.6 & 49.8 & \textbf{59.3}\\
\bottomrule
\end{tabular}
\caption{Performance of pre-trained (PT) and instruction fine-tuned (IT) models on long context benchmarks at different context lengths.}
\label{tab:pt_fs2}
\end{table}

\subsection{Performance of IT models}

\begin{table}[h!]
\centering
\begin{tabular}{@{}l  ccc @{}}
\toprule
&  4B & 12B & 27B \\
\midrule
MMMU (val)                & 48.8 & 59.6  & \textbf{64.9} \\
DocVQA              & 75.8 & \textbf{87.1}  & 86.6 \\
InfoVQA             & 50.0 & 64.9  & \textbf{70.6} \\
TextVQA             & 57.8 & \textbf{67.7}  & 65.1 \\
AI2D                & 74.8 & 84.2  & \textbf{84.5} \\
ChartQA              & 68.8 & 75.7  & \textbf{78.0} \\
VQAv2 (val)               & 62.4 & \textbf{71.6}  & 71.0 \\
MathVista (testmini)           & 50.0 & 62.9  & \textbf{67.6} \\
\bottomrule
\end{tabular}
\caption{Performance of instruction fine-tuned (IT) models on multimodal benchmarks. If not mentioned, these results are on the final test set of each dataset with P\&S applied.}
\label{tab:it_mm}
\end{table}

We report in Table~\ref{tab:it_fs2}, additional benchmarks on our IT models.
Note that N2C refers to Natural2Code, the Gemini 1.0 internal held-out dataset, which uses author-generated sources instead of web-based information.
BBEH refers to BIG-Bench Extra Hard~\citep{kazemi2025big}, a challenging LLM reasoning benchmark that aggregates several reasoning tasks~\citep{kazemi2024boardgameqa, nie2024moca,kiciman2023causal,tyen2023llms,kazemi2023geomverse,sanchez2024linguini,hessel2022androids,zhang2024humor,yamada2023evaluating,fatemi2024test,white2024livebench,shah2024causal}. ECLeKTic refers to~\cite{goldman2025eclekticnovelchallengeset}. We report the micro average score. More evaluation details are described in Table~\ref{tab:eval_detail_it}.

\subsection{Performance of IT models on video understanding}

\begin{table}[h!]
\centering
\begin{tabular}{@{}l  ccc @{}}
\toprule
&  4B & 12B & 27B \\
\midrule
Perception Test MCVQA & 50.6 & 54.9  & 58.1 \\
ActivityNet-QA  & 46.3 & 50.4  & 52.8 \\
\bottomrule
\end{tabular}
\caption{Performance of instruction fine-tuned (IT) models on vision understanding benchmarks using 0 shot with 16 frames linspace. Perception Test consists of real-world videos designed to show perceptually interesting situations and we report results on the multiple choice video QA benchmark in terms of top-1 accuracy. ActivityNet-QA reports standard gpt-evaluation. }
\label{tab:it_mm_video}
\end{table}

\begin{table*}[t]
\centering
\setlength{\tabcolsep}{1.0em}
\begin{tabular}{@{}l ccc c cccc @{}}
\toprule
&\multicolumn{3}{c}{Gemma~2} && \multicolumn{4}{c}{Gemma 3} \\ 
\cmidrule{2-4}
\cmidrule{6-9}
& 2B &  9B & 27B  && 1B & 4B & 12B & 27B \\
\midrule
MMLU & 56.1 & 71.3 & 76.2 && 38.8 & 58.1 & 71.9 & \textbf{76.9} \\
MBPP & 36.6 & 59.2 & 67.4 && 35.2 & 63.2 & 73.0 & \textbf{74.4} \\
HumanEval & 20.1 & 40.2 & 51.8 && 41.5 & 71.3 & 85.4 & \textbf{87.8} \\
N2C & 46.8 & 68.3 & 77.3 && 56.0 & 70.3 & 80.7 & \textbf{84.5} \\
LiveCodeBench & 7.0 & 20.0 & 29.0 && 5.0 & 23.0 & 32.0 & \textbf{39.0} \\
GSM8K & 62.6 & 88.1 & 91.1 && 62.8 & 89.2 & 94.4 & \textbf{95.9} \\
MATH & 27.2 & 49.4 & 55.6 && 48.0 & 75.6 & 83.8 & \textbf{89.0} \\
HiddenMath & 2.0 & 8.0 & 12.0 && 15.0 & 42.0 & 51.0 & \textbf{56.0} \\
BBH & 41.4 & 69.0 & 74.9 && 39.1 & 72.2 & 85.7 & \textbf{87.6} \\
BBEH & 5.9 & 9.8 & 14.8 && 7.2 & 11.0 & 16.3 & \textbf{19.3} \\
IFEval & 80.4 & 88.4 & \textbf{91.1} && 80.2 & 90.2 & 88.9 & 90.4 \\
\midrule
GMMLU-Lite & 41.9 & 64.8 & 68.6 && 34.2 & 54.5 & 69.5 &  \textbf{75.1}\\
ECLeKTic    &  5.3 & 11.8 & \textbf{17.6} && 1.4 & 4.6 & 10.3 & 16.7 \\
\wmtpp      &  37.4 & 48.7 & 51.7 && 35.9 & 46.8 & 51.6 & \textbf{53.4} \\
\bottomrule
\end{tabular}
\caption{Performance of instruction fine-tuned (IT) models of different sizes on more internal and external benchmarks.}
\label{tab:it_fs2}
\end{table*}

\noindent\textbf{Additional multimodal evaluations.}
Gemma 3 IT models were evaluated on common vision benchmarks following the evaluation protocol of Gemini 1.5 \citep{geminiteam2024gemini}. The results are given in Table~\ref{tab:it_mm} when P\&S is activated.

\begin{table*}[t]
\centering
\begin{tabular}{l c c c c c}
\toprule
\textbf{Evaluation}        & \textbf{Metric} & \textbf{Type} & \textbf{n-shot} & \textbf{COT} & \textbf{Norm} \\ 
\midrule
\textbf{MBPP}              & pass@1                  & sampling   & 3-shot    &         &                  \\ 
\textbf{HumanEval}         & pass@1                  & sampling   & 0-shot    &         &                  \\ 
\textbf{HellaSwag}         & Accuracy                & scoring    & 10-shot   &         & Char-Len         \\ 
\textbf{BoolQ}             & Accuracy                & scoring    & 0-shot    &         & Char-Len         \\ 
\textbf{PIQA}              & Accuracy                & scoring    & 0-shot    &         & Char-Len         \\ 
\textbf{SIQA}              & Accuracy                & scoring    & 0-shot    &         & Char-Len         \\ 
\textbf{TriviaQA}          & Accuracy                & sampling   & 5-shot    &         &                  \\ 
\textbf{Natural Questions} & Accuracy                & sampling   & 5-shot    &         &                  \\ 
\textbf{ARC-C}             & Accuracy                & scoring    & 25-shot   &         & Char-Len         \\ 
\textbf{ARC-E}             & Accuracy                & scoring    & 0-shot    &         & Char-Len         \\ 
\textbf{WinoGrande}        & Accuracy                & scoring    & 5-shot    &         & Char-Len         \\ 
\textbf{BBH}               & Accuracy                & sampling   & few-shot  & Yes     &                  \\ 
\textbf{DROP}              & Token F1 score          & sampling   & 1-shot    &         &                  \\ 
\textbf{AGIEval}           & Accuracy                & sampling   & 3-5-shot  &         &                  \\ 
\textbf{MMLU}              & Accuracy                & scoring    & 5-shot    &         & Char-Len         \\ 
\textbf{MATH}              & Accuracy                & sampling   & 4-shot    & Yes     &                  \\ 
\textbf{GSM8K}             & Accuracy                & sampling   & 8-shot    & Yes     &                  \\ 
\textbf{GPQA Diamond}              & Accuracy                & sampling   & 5-shot    & Yes     &                  \\ 
\textbf{MMLU-Pro}          & Accuracy                & sampling   & 5-shot    & Yes     &                  \\ 
\textbf{MGSM}              & Accuracy                & sampling   & 8-shot    &         &                  \\ 
\textbf{FLoRes}            & CHaRacter-level F-score & sampling   & 1-shot    &         &                  \\ 
\textbf{Global-MMLU-Lite}  & Accuracy                & scoring    & 5-shot    &         & Char-Len         \\ 
\textbf{XQuAD}             & CHaRacter-level F-score & sampling   & 5-shot    &         &                  \\ 
\textbf{WMT24++}           & CHaRacter-level F-score & sampling   & 5-shot    &         &                  \\ 
\textbf{ECLeKTic}          & ECLeKTic score          & sampling   & 2-shot    &         & First-line/strip \\ 
\textbf{XQuAD Indic}       & CHaRacter-level F-score & sampling   & 5-shot    &         &                  \\ 
\textbf{XOR QA IN-EN}      & CHaRacter-level F-score & sampling   & 5-shot    &         &                  \\
\textbf{XOR QA IN-XX}      & CHaRacter-level F-score & sampling   & 5-shot    &         &                  \\ 
\textbf{FLoRes Indic}      & CHaRacter-level F-score & sampling   & 5-shot    &         &                  \\ 
\textbf{RULER}             & Accuracy                & sampling   & 0-shot    &         &                  \\ 
\textbf{MRCR}              & MRCR score       & sampling   & few-shot    &         &                  \\
\bottomrule
\end{tabular}
\caption{Details on text benchmarks. Char-Len stands for Character Length Normalization and COT stands for Chain-Of-Thought prompting.}
\label{tab:eval_detail_text}
\end{table*}

\begin{table*}[t]
\centering
\begin{tabular}{l c c c}
\toprule
\textbf{Evaluation}        & \textbf{Metric} & \textbf{Type} & \textbf{n-shot} \\ 
\midrule
\textbf{COCO Caption}      & Cider score     & sampling      & 4-shot           \\
\textbf{DocVQA}            & ANLS score      & sampling      & 4-shot           \\
\textbf{InfographicVQA}    & ANLS score      & sampling      & 4-shot           \\
\textbf{MMMU}              & Accuracy        & sampling      & 3-shot text only \\
\textbf{TextVQA}           & Accuracy        & sampling      & 4-shot           \\
\textbf{RealWorldQA}       & Accuracy        & sampling      & 4-shot text only \\
\textbf{ReMI}              & Accuracy        & sampling      & 4-shot           \\
\textbf{AI2D}              & Accuracy        & sampling      & 4-shot           \\
\textbf{ChartQA}           & Accuracy        & sampling      & 4-shot           \\
\textbf{VQA v2}            & Accuracy        & sampling      & 4-shot           \\
\textbf{BLINK}             & Accuracy        & sampling      & 0-shot           \\
\textbf{OK-VQA}            & Accuracy        & sampling      & 4-shot           \\
\textbf{TallyQA}           & Accuracy        & sampling      & 4-shot         \\
\textbf{SpatialSense VQA}  & Accuracy        & sampling      & 4-shot           \\
\textbf{CountBench VQA}    & Accuracy        & sampling      & 0-shot           \\
\bottomrule
\end{tabular}
\caption{Details on vision benchmarks. No Chain-Of-Thought prompting nor normalization.}
\label{tab:eval_detail_mm}
\end{table*}

\begin{table*}[t]
\centering
\begin{tabular}{l c c c c}
\toprule
\textbf{Evaluation}        & \textbf{Metric} & \textbf{Type} & \textbf{n-shot}  & \textbf{COT}\\ 
\midrule
\textbf{MMLU}           &   Accuracy                & sampling  &  0-shot  &       \\           
\textbf{MBPP}           &   pass@1                  & sampling  &  3-shot  &       \\           
\textbf{HumanEval}      &   pass@1                  & sampling  &  0-shot  &       \\           
\textbf{N2C}            &   pass@1                  & sampling  &  0-shot  &       \\           
\textbf{LiveCodeBench}  &   Average over 8 samples  & sampling  &  0-shot  &  Yes  \\              
\textbf{GSM8K}          &   Accuracy                & sampling  &  0-shot  &  Yes  \\   
\textbf{GPQA Diamond}    & Accuracy                & sampling   & 0-shot    & Yes    \\ 
\textbf{MATH}           &   Accuracy                & sampling  &  0-shot  &       \\           
\textbf{HiddenMath}     &   Accuracy                & sampling  &  0-shot  &       \\           
\textbf{BBH}            &   Accuracy                & sampling  &  0-shot  &       \\           
\textbf{BBEH}           &   Accuracy                & sampling  &  0-shot  &       \\           
\textbf{IFEval}         &   Accuracy                & sampling  &  0-shot  &       \\           
\textbf{Global-MMLU-lite}    &   Accuracy                & sampling  &  0-shot  &  Yes  \\              
\textbf{ECLeKTic}       &   ECLeKTic score          & sampling  &  0-shot  &       \\           
\textbf{WMT24++}        &   CHaRacter-level F-score & sampling  &  0-shot  &       \\  
\bottomrule
\end{tabular}
\caption{Details on instruction fine-tuned (IT) benchmarks. No normalization.}
\label{tab:eval_detail_it}
\end{table*}

\end{document}